\documentclass[sigconf]{acmart}

\AtBeginDocument{%
  \providecommand\BibTeX{{%
    \normalfont B\kern-0.5em{\scshape i\kern-0.25em b}\kern-0.8em\TeX}}}

\copyrightyear{2024} 
\acmYear{2024} 
\setcopyright{rightsretained} 
\acmConference[SIGSPATIAL '24]{The 32nd ACM International Conference on
Advances in Geographic Information Systems}{October 29-November 1,2024}{Atlanta, GA, USA}
\acmBooktitle{The 32nd ACM International Conference on Advances in
Geographic Information Systems (SIGSPATIAL '24), October 29-November 1,
2024, Atlanta, GA, USA}
\acmDOI{10.1145/3678717.3691254}
\acmISBN{979-8-4007-1107-7/24/10}

\usepackage{multirow}
\usepackage[normalem]{ulem}
\useunder{\uline}{\ul}{}
\usepackage{bbm}
\usepackage{enumitem}
\usepackage{tablefootnote}
\usepackage{makecell}
\usepackage[ruled,linesnumbered,longend]{algorithm2e}
\usepackage{algpseudocode}  
\usepackage{amsmath}

\begin{document}

\title{Large-scale Urban Facility Location Selection with Knowledge-informed Reinforcement Learning}


\author{Hongyuan Su\footnotemark[1], Yu Zheng}
\authornote{Both authors contributed equally to this research.}
\affiliation{%
  \institution{Department of Electronic Engineering, BNRist, 
  \\Tsinghua University}
  \city{Beijing}
  \country{China}
}

\author{Jingtao Ding}
\affiliation{%
  \institution{Department of Electronic Engineering, BNRist, 
  \\Tsinghua University}
  \city{Beijing}
  \country{China}
}

\author{Depeng Jin}
\affiliation{%
  \institution{Department of Electronic Engineering, BNRist,
  \\Tsinghua University}
  \city{Beijing}
  \country{China}
}

\author{Yong Li}
\authornote{Corresponding author (liyong07@tsinghua.edu.cn).}
\affiliation{%
  \institution{Department of Electronic Engineering, BNRist, 
  \\Tsinghua University}
  \city{Beijing}
  \country{China}
}

\renewcommand{\shortauthors}{Su and Zheng, et al.}

\begin{abstract}
The facility location problem (FLP) is a classical combinatorial optimization challenge aimed at strategically laying out facilities to maximize their accessibility.
In this paper, we propose a reinforcement learning method tailored to solve large-scale urban FLP, capable of producing near-optimal solutions at superfast inference speed.
We distill the essential swap operation from local search, and simulate it by intelligently selecting edges on a graph of urban regions, guided by a knowledge-informed graph neural network, thus sidestepping the need for heavy computation of local search.
Extensive experiments on four US cities with different geospatial conditions demonstrate that our approach can achieve comparable performance to commercial solvers with less than 5\% accessibility loss, while displaying up to 1000 times speedup.
We deploy our model as an online geospatial application at \textcolor{blue}{\url{https://huggingface.co/spaces/randommmm/MFLP}}.
\end{abstract}


\keywords{Facility location problem, reinforcement learning, large-scale optimization}

\maketitle

\section{Introduction}
In real cities, facility layout tends to deviate from residential demands for corresponding services, leading to costly travel~\cite{xu2020deconstructing,zheng2023spatial}.
Therefore, optimizing accessibility by strategically locating urban facilities is crucial for creating more sustainable and inclusive cities.
In fact, facility location problem (FLP) is a classic combinatorial optimization (CO) problem~\cite{shmoys1997approximation,drezner2004facility}, which is notoriously challenging due to the NP-hardness inherent in selecting $p$ urban regions to place facilities from $N$ candidate regions~\cite{maranzana1964location}.
As both $N$ and $p$ are typically large in urban contexts, designing a reliable algorithm that delivers satisfactory solutions within reasonable timeframes is difficult.
Furthermore, unlike the standard FLP setup, urban facilities exhibit multiple types, each with distinct budgets and  residential needs, further complicating the task.

FLP is traditionally addressed using heuristics, metaheuristics~\cite{kariv1979algorithmic,rolland1997efficient,gwalani2021evaluation} and commercial mixed integer programming (MIP) solvers~\cite{andersen2000mosek}.
Despite the near-optimal solutions they offer, the heavy computation of local search inherent in these approaches renders them suitable only for small-scale problems, while they tend to be unacceptably slow~\cite{wang2022towards}.
On the other hand, machine learning models are proposed to tackle FLP, \textit{i.e.,} ML4CO~\cite{vinyals2015pointer,su2024rumor,zheng2023road}, which replace intensive local search with learnable neural networks.
Though ML4CO significantly accelerates solution generation through fast model inference, it has drastically degenerated accessibility performance for the overlooking rich prior knowledge inherent in FLP.

In this paper, we present a deep reinforcement learning (DRL) method to solve the large-scale urban facility location problem, which is able to achieve comparable accessibility performance to commercial solvers while generating solutions with up to 1000 times speedup.
Specifically, we construct an urban graph with geospatial information, based on which we distill the essential \textit{swap} operation in local search, replacing inefficient exploration with rapid edge selections which are informed by a GNN tailored to the FLP. 
Additionally, we integrate tabu search~\cite{rolland1997efficient} and node benefits~\cite{resende2007fast} into the graph via dynamic wiring to avoid redundant and low-quality solutions.
Finally, to address practical urban FLP with multiple types of facilities, we design a \textit{divide-and-conquer} algorithm to solve the problem with two stages, which first accomplishes single facility solutions independently, then conducts adjustments, both using the unified \textit{SWAP} operation and GNN model.

To summarize, the contributions of this paper are as follows,
\begin{itemize}[leftmargin=*]
    \item To the best of our knowledge, we are the first to investigate practical large-scale urban FLP with multiple types of facilities.
    \item We develop a knowledge-informed RL method that incorporates rich prior knowledge of FLP into the design of Markov Decision Process (MDP) and GNN, as well as a divide-and-conquer algorithm that solves the problem with the unified RL model, bypassing the need for heavy computation of local search.
    \item We conduct experiments using real-world large-scale urban geospatial and human mobility data, demonstrating that our approach can achieve comparable performance to commercial solvers with less than 5\% performance loss while generating solutions up to 1000 times faster.
\end{itemize}
\section{Problem Formulation}\label{sec::problem}


\begin{figure}[t]
    \centering
    \includegraphics[width=0.95\linewidth]{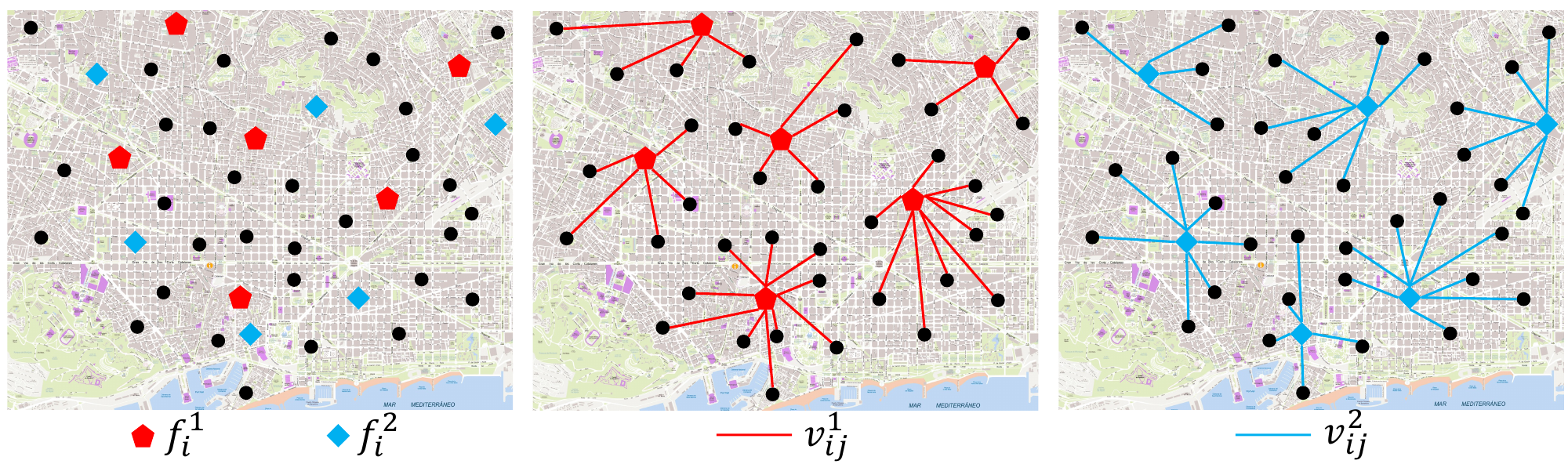}
    \vspace{-10px}
    \caption{An example of MFLP.
    Red pentagons and blue diamonds are the placed two types of facilities.
    Red and blue edges indicate that residents of each region visit their closest facilities to access basic services.
    Best viewed in color.}
    \label{fig:prob_form}
    \vspace{-12px}
\end{figure}


Given the total number of facilities $P$, candidate locations for facilities $\mathcal{N}=\{N_1,N_2,\cdots,N_n\}$, and corresponding demands $\mathcal{H}=\{h_1,h_2,\cdots,h_n\}$, the FLP aims to maximize the accessibility of facilities.
Different from the FLP, which focuses on a single type of facility, we investigate Multi-Facility Location Problem (MFLP), which involves $\mathcal{K}$ different types of facilities which are distinguished by different superscripts, and the type $k$ facility has to be placed in $P^k$ different regions among $\mathcal{N}$, with the corresponding residential demands denoted as $\mathcal{H}^k = \{h_1^k, \ldots, h_n^k\}$.
The MFLP aims to optimize accessibility to all types of facilities, which can be measured by the access cost (AC) as follows:
\begin{equation}\label{eq:ac}
AC =\sum_{k\in\mathcal{K}}\sum_{i=1}^n\sum_{j=1}^n h_{i}^kd_{ij}Y_{ij}^k,
\end{equation}
where $d_{ij}$ denotes the distance between location $N_i$ and $N_j$, and $Y_{ij}$ indicates whether demand $h_i$ at $N_i$ is satisfied by a facility at $N_j$.

For the purpose of modeling the interactions between different facilities aforementioned, we introduce incompatibility to MFLP, which is the key to differentiating MFLP from multiple independent FLPs.
Incompatibility requires only one specific type of facility can be placed at each region, and can be characterized as,
\begin{equation}
\label{eq:incompatibility}
    X_i^p X_i^q\equiv 0, \,\forall p,q\in \mathcal{K}, p\neq q,N_i \in \mathcal{N},
\end{equation}
where $X_i^p$ is an indicator variable for whether a facility of type $p$ is placed at $N_i$.

Figure \ref{fig:prob_form} provides an illustrative example of MFLP for two different types of urban facilities.
In particular, when $|K|=1$, the MFLP will degenerate into the standard FLP problem.





\section{Method}


We introduce a knowledge-informed DRL approach to solve the large-scale FLP.
Inspired by local search methods, the SWAP operator is developed to fully leverage the searched solution space.
Based on the SWAP operator, we define a swap graph for the large-scale FLP and formulate the FLP as a sequential edge selection problem with a domain knowledge-driven dynamic wiring to identify high-quality SWAP operations.
Additionally, a divide-and-conquer framework is further proposed to address the real-world MFLP, taking into account constraints between different types of facilities with a two-stage process.


\subsection{The Universal SWAP operator}\label{sec:swap}
Cities of large scale contain thousands of candidate regions and require hundreds of facilities, making it time-consuming to generate solutions from scratch due to its repeated computations.
To make full use of the already obtained solutions, we distill the universal SWAP operator based on the idea of local search~\cite{rolland1997efficient}, which generates new solutions by fine-tuning and improving the current solution.
Specifically, each SWAP operator takes two locations, \textit{insert} and \textit{remove}, and moves the facility at \textit{remove} to \textit{insert}.
Through successive SWAP operators, the initial solution can be adjusted towards better solutions iteratively until reaching the optimal.

\begin{figure*}[t]
    \centering
    \includegraphics[width=0.99\linewidth]{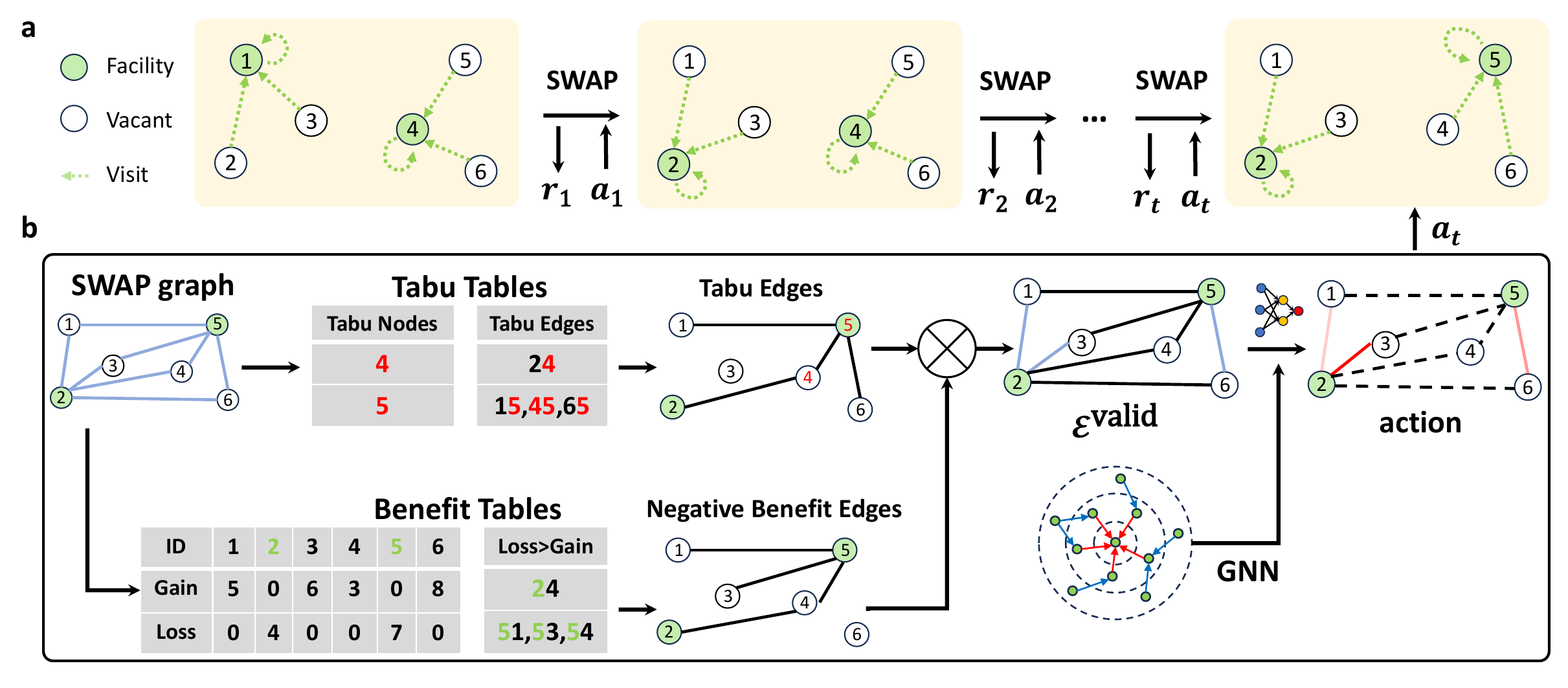}
    \vspace{-10px}
    \caption{The overall RL framework solving the large-scale FLP.
    (a) FLP as a sequential decision-making problem on the swap graph, where one single facility is moved from its origin region (green) to one of the vacant regions (white) through the SWAP operator at each step.
    (b) The knowledge-informed RL agent for edge selection leverages the dynamic wiring technique to filter out both Tabu Edges and Negative Benefit Edges.
    Best viewed in color.}
    \label{fig:RL}
    \vspace{-8px}
\end{figure*}

\subsection{Knowledge-Informed RL}\label{sec:rl}
As introduced previously, heuristic methods are accurate but slow, while ML4CO approaches accelerate solution generation with significant accuracy deterioration.
We propose a knowledge-informed RL method that uses the best of both worlds, generating accurate solutions at a superfast speed, as illustrated in Figure \ref{fig:RL}.

\noindent\textbf{MDP on Graph.}
We regard all the candidate locations as nodes and add edges between nodes with and without facility to construct a swap graph, which can be formulated as follows,
\begin{equation}\label{eq:swapg}
    \mathcal{G}_t = \{\mathcal{N},\mathcal{E}_t\}\,, e_{ij,t}=\mathbbm{1}\{X_{i,t} < X_{j,t}\}.
\end{equation}
The SWAP operator is thus equivalent to edge selection on the swap graph, as illustrated in Figure \ref{fig:RL} (b).
Moreover, since the solution of MFLP is generated through a sequence of SWAP operators, solving MFLP can be framed as an MDP with the following elements:
\begin{itemize}[leftmargin=*]
    \item \textbf{State} summarizes the information about the topology of urban regions and the placement of facilities.
    \item \textbf{Action} denotes choosing an edge on the swap graph and exchanging the corresponding facility placement.
    \item \textbf{Reward} is the reduction of access cost defined by Equation (\ref{eq:ac}).
\end{itemize}

\noindent\textbf{Dynamic Wiring by FLP Knowledge.}
Not all edges on the swap graph benefit the reduction of the access cost, particularly when the solution is close to the optimal or falls into a local optimum.
Here, we introduce a dynamic wiring technique informed by FLP knowledge, to initially assess the exploration value of each action.
Specifically, as shown in Figure \ref{fig:RL}, our dynamic wiring involves filtering two categories of edges that contribute less to the optimal solution.
\noindent\underline{\textit{Tabu Edges}} are inspired by Tabu Search~\cite{rolland1997efficient}, and recently visited locations by SWAP operators will be recorded as Tabu nodes, and edges containing Tabu Nodes are considered as Tabu Edges.
\noindent\underline{\textit{Negative Benefit Edges}} indicate actions with significant negative returns.
The change of total access cost after taking action can be divided into three parts: the \textit{gain} if only placing facility, the \textit{loss} if only removing facility, and the \textit{extra} term caused by simultaneously placing and removing.
Notably, if two locations are far enough apart, the \textit{extra} always equals to zero~\cite{resende2007fast}.
Consequently, if locations are far apart and the \textit{gain} is less than the \textit{loss}, the SWAP operator will increase the total access cost instead.
Based on the above FLP knowledge, the dynamic wiring technique highlights the quality actions for facility placement and guarantees the accessibility performance of the obtained solutions.

\noindent\textbf{GNN State Encoder.}
We design a GNN-based model to conduct SWAP operations, bypassing the time-consuming local search.
Specifically, the proposed GNN considers both local topological relationship and global facility allocations to learn comprehensive embeddings for high-quality edges on the swap graph as follows,

\begin{align}
    &\mathbf{n}_i^{0} = \mathbf{W}_n^{0}\mathbf{A}_{{n}_i},\\
    &\mathbf{n}_{i}^{l+1} = \mathbf{n}_{i}^{l} + \tanh(\sum_{e_{ij}} \mathbf{W}_{n}^{l+1}\mathbf{n}_j^{l}),\\
     &\mathbf{e}_{ij} = [ \mathbf{n}_i^{L} \,\|\, \mathbf{n}_j^{L} ],
\end{align}
where $\mathbf{A}_{{N}_i^k}$ is the input attributes for nodes when placing facility $k$, $\mathbf{W}_{n}^{l}$ is the weight matrix of the $l$-th layer of the GNN, $\mathbf{e}_{ij}$ is the representation of edges on the swap graph, $\|$ means concentration, and $\mathbf{n}_i^{L}$ is the output of the last layer and $L$ is a hyper-parameter for the number of GNN layers.

\subsection{Divide-and-Conquer Framework}
The proposed RL framework solves FLP with superior efficiency.
We now extend it to address the incompatibility in the MFLP with a divide-and-conquer framework, which consists of two successive stages.
In \textbf{Stage \uppercase\expandafter{\romannumeral1}}, the agent searches for the optimal locations for each type of facility using the SWAP operator independently, starting from a greedily constructed initial solution.
Since these independent solutions may conflict with the incompatibility constraints described by Equation (\ref{eq:incompatibility}), in \textbf{Stage \uppercase\expandafter{\romannumeral2}} the agent further refines these independent solutions by eliminating the conflicts via the SWAP operators as well.
Stage \uppercase\expandafter{\romannumeral2} continues until all incompatibility conflicts are resolved.

With this divide-and-conquer framework, we can reuse the proposed RL algorithm for large-scale FLP to address the MFLP.
On the one hand, both of the two stages share the aligned objective of minimizing the access cost, \textit{i.e.}, generating separate optimal placements for each type of facility with lowest access cost in Stage \uppercase\expandafter{\romannumeral1} and resolving incompatibilities with the minimal increase in access cost in Stage \uppercase\expandafter{\romannumeral2}.
On the other hand, we accomplish both stages employing the same universal SWAP operator.
Consequently, our knowledge-informed RL approach kills two birds with one stone by intelligently conducting SWAP operators to optimize access cost, unifying the two challenging problems of FLP and MFLP.

\section{Experiments}

We conduct extensive experiments based on the real world urban geospatial data and human mobility data from \href{https://www.safegraph.com}{SafeGraph}.
Our method is compared against traditional methods such as genetic algorithm (GA), tabu search (TS), local search (LS).
Moreover, advanced machine learning approach for CO problem (CardNN)~\cite{wang2022towards} and commercial solver Gurobi are included.

\begin{table}[t]
\centering
\caption{
Performance on MFLP, evaluated by access cost (AC) and algorithm runtime (RT) in seconds.
Nan means the method fails to solve within 24h or with 100GB RAM. 
}
\vspace{-10px}
\tabcolsep=0.09cm
\begin{tabular}{c|cc|cc|cc}
\toprule
\multirow{2}{*}{\textbf{Method}} 
& \multicolumn{2}{c|}{\textbf{New York}} &\multicolumn{2}{c|}{\textbf{Los Angeles}} &\multicolumn{2}{c}{\textbf{Chicago}}\\
                                 & \textbf{AC}$\downarrow$ & \textbf{RT}$\downarrow$ & \textbf{AC}$\downarrow$ & \textbf{RT}$\downarrow$ & \textbf{AC}$\downarrow$ & \textbf{RT}$\downarrow$\\
\midrule
GA & 31.04 & \underline{81.25} & 25.99 & 123.87 & 30.86 & \underline{89.53}\\
    TS & 28.86 & 3489.91 & 23.91 & 2714.75 & 28.16 & 3049.60\\
LS & 27.10 & 311.54 & 23.63 & 498.35 & 28.93 & 286.62\\
VNS & 26.94 & 1591.36 & 22.13 & 1189.88 & 26.71 & 1046.94\\
CardNN & \underline{26.62} & 906.29 & \underline{21.76} & 784.11 & 26.72 & 609.28\\
Gurobi (120s) & 27.48 & 120 & 23.43 & \underline{120} & 28.04 & 120\\
Gurobi (optimal) & nan & nan & nan & nan & nan & nan\\
RSSV & 26.88 & 925.84 & 22.55 & 879.73 & \underline{26.29} & 848.02\\
DRL-GNN (ours) & \textbf{26.29} & \textbf{36.62} & \textbf{21.08} & \textbf{24.65} & \textbf{25.58} & \textbf{26.86}\\
\hline
impr\% / speed up & +4.9 & 24.8 & +4.2 & 31.8 & +4.7 & 39.0\\
\toprule
\end{tabular}
\renewcommand{\arraystretch}{1.0}
\vspace{-12px}
\label{tab:overall}
\end{table}

\subsection{Overall Performance}
We evaluate our model across different large-scale cities with the results illustrated in Table \ref{tab:overall}, from which we have the following observations,
\begin{itemize}[leftmargin=*]
    \item \textbf{Heuristic methods are ineffective for large-scale MFLP.}
    Classic heuristic methods like TS, LS, and VNS fail to generate high-quality solutions for MFLP, showing a noticeable relative gap of nearly 10\% compared to the optimal solutions.
    In addition to the worse accessibility, the runtime is also significantly longer than other methods.
    These observations indicate the need for further improvements in search effectiveness.
    \item \textbf{Our methods balances effectiveness and efficiency as well.}
    Aside from Gurobi (optimal), our approach achieves the lowest access cost and offers the fastest solution generation speed across all four cities. 
    Although the access cost is 4.2\% higher than Gurobi (optimal) in the medium-sized Los Angeles scenario, the solution is generated 32 times faster.
    Moreover, our approach outperforms other ML4CO methods like CardNN, with a 2.5\% lower access cost and a nearly 87-fold speed increase.
\end{itemize}

Since the complexity of the MFLP is much more higher than that of FLP, Gurobi often fails on solving large-scale MFLP.
Considering that the performance of solving FLP serves as the foundation for MFLP, we conduct an additional comparison of different methods on the FLP, where Gurobi (optimal) can offer the optimal solutions.
To investigate the performance on problems of different scales, we use synthetic data of urban regions and facility access needs, then evaluate the solutions generated by different methods.

\begin{figure}[t]
    \centering
    \includegraphics[width=1\linewidth]{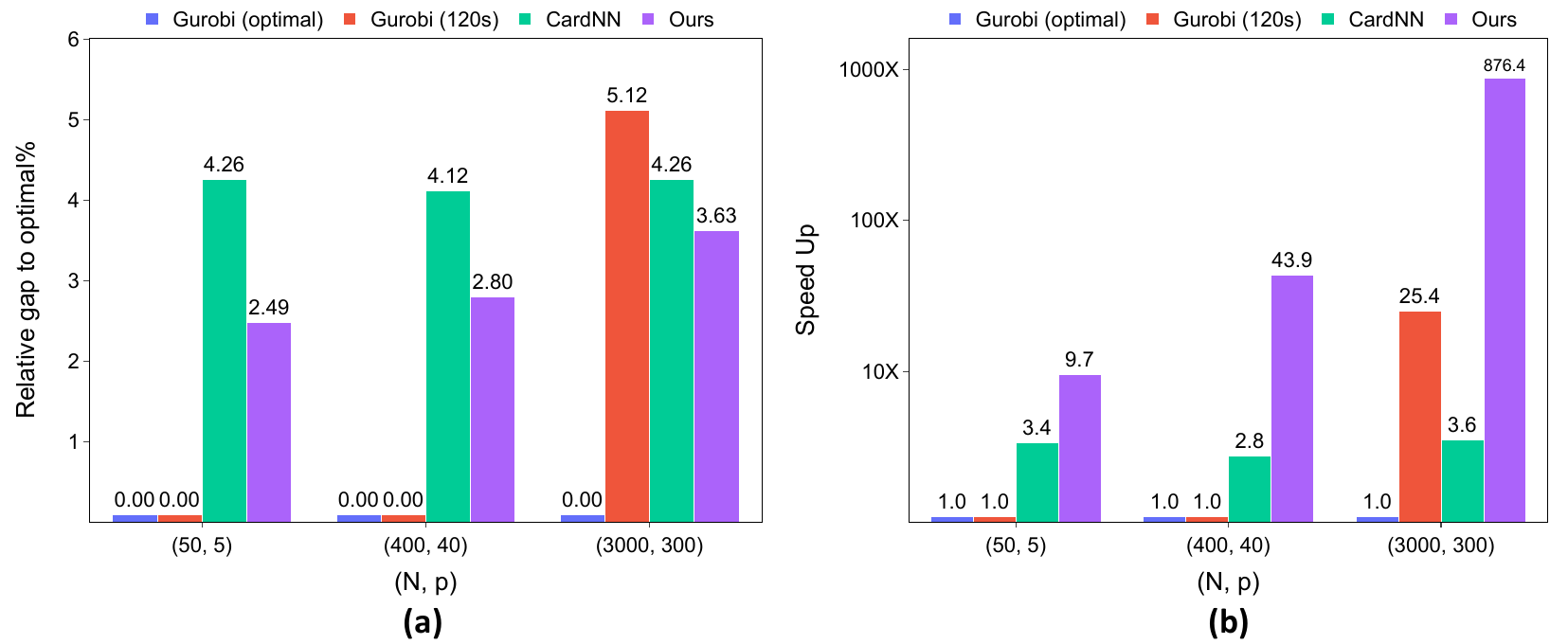}
    \vspace{-20px}
    \caption{FLP performance comparison.
    For each pair (N,p), we generate 1000 simulated cities and calculate 
    (a) the average relative gap\% between each method and Gurobi (optimal).
    (b) the average speed up compared to Gurobi (optimal).}
    \label{fig:np}
    \vspace{-10px}
\end{figure}

As illustrated in Figure \ref{fig:np}, our method excels in facility placement at remarkable speed across problems of different scales.
Specifically, the relative gap of access cost between our approach and the optimal solutions remains below 3.7\%.
Notably, the speedup of our method reaches over 876 times when the city is huge and has numerous facilities, such as $(N,p)=(3000,300)$, highlighting its outstanding performance in both solution quality and algorithm runtime.

\section{Conclusion}


In this paper, we present a knowledge-informed RL method tailored to solve real-world FLP and MFLP. 
By infusing rich prior knowledge into the dynamical wiring of GNN, our application can produce near-optimal solutions for large-scale FLP with up to 1000 times speedup compared to commercial solvers.
Particularly, for New York city-scale scenarios, which are too large such that existing commercial solvers fail to handle, our method generates reliable solutions within 40 seconds. 
Our method also enables a seamless trade-off between the accessibility performance and solution generation time through adjusting the number of swap steps performed.


\begin{acks}
This work is supported in part by National Key Research and Development Program of China under grant 2022ZD0116402 and National Natural Science Foundation of China under grant U22B2057 and 62272260.
This work is also supported in part by Tsinghua University-Toyota Research Center and Beijing National Research Center for Information Science and Technology (BNRist).
\end{acks}

\bibliographystyle{ACM-Reference-Format}


\end{document}